# Robust Multiple Manifolds Structure Learning

**Dian Gong, Xuemei Zhao, Gérard Medioni**　　　　　　　　　　　　　DIANGONG,XUEMEIZ,MEDIONI@USC.EDU

University of Southern California, Institute for Robotics and Intelligent Systems, Los Angeles, CA 90089, USA

## Abstract

We present a robust multiple manifolds structure learning (RMMSL) scheme to robustly estimate data structures under the multiple low intrinsic dimensional manifolds assumption. In the local learning stage, RMMSL efficiently estimates local tangent space by weighted low-rank matrix factorization. In the global learning stage, we propose a robust manifold clustering method based on local structure learning results. The proposed clustering method is designed to get the flattest manifolds clusters by introducing a novel curved-level similarity function. Our approach is evaluated and compared to state-of-the-art methods on synthetic data, handwritten digit images, human motion capture data and motor-bike videos. We demonstrate the effectiveness of the proposed approach, which yields higher clustering accuracy, and produces promising results for challenging tasks of human motion segmentation and motion flow learning from videos.

## 1. Introduction

The concept of manifold has been extensively used in almost all aspects of machine learning such as non-linear dimension reduction (Saul & Roweis, 2003), visualization (Lawrence, 2005; van der Maaten, 2009), semi-supervised learning (Singh et al., 2009), multi-task learning (Agarwal et al., 2010) and regression (Steinke & Hein, 2009). Related methods have been applied to many real-world problems in computer vision, computer graphics, web data mining and more. Despite the success of manifold learning (in this paper, manifold learning refers to any learning technique that explicitly assumes data have manifolds structures), people find there are several fundamental challenges in real applications,

(1) Multiple manifolds with possible intersections: in many circumstances there is no unique (global) manifold but a number of manifolds with possible intersections. For instance, in handwritten digit images, each digit forms its own manifold in the observed feature space. For human motion, joint-position (angle) of key points in body skeleton form low dimensional manifolds for each specific action (Urtasun et al., 2008). In these situations, modeling data as a union of (linear or non-linear) manifolds rather than a single one can give us a better foundation for many tasks such as semi-supervised learning (Goldberg et al., 2009) and denoising (Hein & Maier, 2007).

(2) Noise and outliers: one critical issue of manifold learning is whether the method is robust to noise and outliers. This has been pointed out in the pointer work on nonlinear dimension reduction (Saul & Roweis, 2003). For nonlinear manifolds, since it is not possible to leverage all data to estimate local data structure, more samples from manifolds are required as the noise level increases.

(3) High curvature and local linearity assumption: typical manifold learning algorithms approximate manifolds by the union of locally linear patches (possibly overlapped). These local patches are estimated by linear methods such as principal component analysis (PCA) and factor analysis (FA) (Teh & Roweis, 2003). However, for high curvature manifolds, many smaller patches are needed, but this often conflicts with the limited number of data samples.

In this paper, we investigate the problem of robustly estimating data structure under the multiple manifolds assumption. In particular, data are assumed to be sampled from multiple smooth submanifolds of (possibly different) low intrinsic dimensionalities with noise and outliers. The proposed scheme named Robust Multiple Manifolds Structure Learning (RMMSL) is composed of two stages. In the local stage, we estimate local manifolds structure taking into account of noise and curvature. The global stage, i.e., manifold clustering and outlier detection, is performed by constructing a multiple kernel similarity graph based on local structure learning results by introducing a novel curved-level similarity function. Thus, issue (1) is explicitly addressed, and (2) and (3) are partially investigated. A demonstration of the proposed approach is given in Fig. 1. It is worth noting that other types of global learning tasks such as dimension reduction, denoising and semi-supervised learning can be handled based on each individual manifold cluster by existing methods (Lawrence, 2005; van der Maaten, 2009; Hein & Maier, 2007; Sinha & Belkin, 2010).

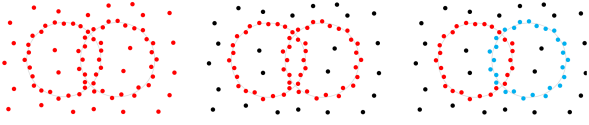

*Figure 1.* A demonstration of RMMSL. From left to right, noisy data points sampled from two intersecting circles with outliers, outlier detection results, and manifolds clustering results after outlier detection.

Our problem statement is rigorously expressed as follows. Data $\{x_i\}_{i=1}^{n_1+n_2} \in \mathbb{R}^{D \times (n_1+n_2)}$ consist of inliers $\{x_i\}$ ($1 \leq i \leq n_1$) and outliers $\{x_j\}$ ($n_1 + 1 \leq j \leq n_1 + n_2$). Inlier points $\{x_i\}_{i=1}^{n_1} \in \mathbb{R}^{D \times n_1}$ are assumed to be sampled from multiple submanifolds $\mathfrak{M}_c$ ($1 \leq c \leq n_c$) as follows,

$$x_i = f_{C(x_i)}(\tau_i) + n_i, i = 1, 2, ..., n_1 \quad (1)$$

where $C(x_i) \in \{1, 2, ..., n_c\}$ is the manifold label function for $x_i$. $n_i$ is inlier noise in $\mathbb{R}^D$. $f_c(\cdot)$ is the smooth mapping function that maps latent variable $\tau_i$ from latent space $\mathbb{R}^{d_c}$ to ambient space $\mathbb{R}^D$ for manifold $\mathfrak{M}_c$. $d_c(< D)$ is the intrinsic dimensionality of $\mathfrak{M}_c$ and it can vary for different manifolds. The task of local structure learning is to estimate the tangent space $T_{x_i}\mathfrak{M}$ and the local intrinsic dimensionality $d_{C(x_i)}$ at each $x_i$. This is equivalent to estimate $J(f_{c_i}(\cdot); \tau_i) \in \mathbb{R}^{D \times d_{c_i}}$, i.e., the Jacobian matrix of $f_{c_i}(\cdot)$ at $\tau_i$. Details are given in section 3. The tasks of global structure learning addressed in this paper are, to detect outliers $\{x_j\}$ ($n_1 + 1 \leq j \leq n_1 + n_2$) and to assign manifolds cluster label $c_i \in \{1, 2, ..., n_c\}$ for inliers $\{x_i\}$ ($1 \leq i \leq n_1$) (Fig. 2). This is given in section 4. Experimental results are shown in section 5 and followed by conclusion in section 6.

## 2. Related Work

Recent works on manifold learning and latent variable modeling focus on low-dimensional embedding and label propagation for high dimensional data, which are usually assumed to be sampled from a single manifold (Saul & Roweis, 2003; Sinha & Belkin, 2010; Lawrence, 2005; van der Maaten, 2009). To address the multiple manifolds problem, (Goldberg et al., 2009) gives theoretical analysis and a practical algorithm for semi-supervised learning with the multi-manifold assumption. As a complementary approach to previous works, RMMSL mainly focuses on unsupervised learning with the multi-manifold assumption and it can be combined with existing approaches[1].

In the global structure learning stage, RMMSL focuses on clustering and outlier detection. Clustering is a long standing problem and we only review manifold related clustering methods. If data lie on a low-dimensional submanifold, distance on the manifold is used to replace the Euclidean distance in the clustering process. This leads to spectral clustering (Ng et al., 2002; Zelnik-Manor & Perona, 2005; Maier et al., 2009), one of the most popular modern clustering algorithms. We refer (von Luxburg, 2007) as a survey for spectral clustering. However, most of these works do not explicitly consider multiple intersecting manifolds.

State-of-the-art multiple subspace learning methods such as (Elhamifar & Vidal, 2009; Vidal et al., 2010; Yan & Pollefeys, 2006) acquire good performance on multiple linear (intersecting) subspace segmentation with the assumption that the intrinsic manifolds have linear structures. However, real data often have nonlinear intrinsic structures, which brings difficulty for them. Nonlinear manifold clustering is investigated in (Goh & Vidal, 2007), which mainly focuses on separated manifolds. As an extension of ISOMAP, (Souvenir & Pless, 2005) proposes an EM algorithm to perform multiple manifolds clustering, but results are sensitive to initializations and the E-step is heuristic. Recently, (Wang et al., 2011) proposes an elegant solution of manifolds clustering by constructing the affinity matrix based on estimated local tangent space.

## 3. Local Manifold Structure Estimation

Correctly and efficiently estimating local data structure is a crucial step for data analysis. As explained before, problems like noise and high curvature make local structure estimation challenging. This section addresses the issue of how to model and represent local structure information on manifolds. We start from local Taylor expansion, and local manifold tangent space is represented by the Jacobian matrix under the local isometry assumption.

**Local Taylor Expansion.** Without additional assumptions, the model in eq. 1 (section 1) is not well defined. For instance, if $f(\cdot)$ and point set $\{\tau_i\}$ satisfy eq. 1, then $f(g^{-1}(\cdot))$ and point set $\{g(\tau_i)\}$ satisfy it too, where $g(\cdot)$ is any invertible and differential mapping function from $\mathbb{R}^d$ to $\mathbb{R}^d$. Thus, the *local isometry* assumption is enforced at point $\tau_i \in \mathbb{R}^d$,

$$||f(\tau) - f(\tau_i)||_2 = ||\tau - \tau_i||_2 + o(||\tau - \tau_i||_2) \quad (2)$$

where $\tau$ is in the $\varepsilon - neighborhood$ of $\tau_i$. The above condition implies that $J(f(\cdot); \tau_i) \in \mathbb{R}^{D \times d}$ is an orthonormal matrix (Smith et al., 2009), i.e., $J^T(f(\cdot); \tau_i) J(f(\cdot); \tau_i) = I_d$.

---
[1] It can also be used to supervised learning with minor modifications.

From eq. 1, by using Taylor expansion to incorporate both Taylor approximation error and inlier noise we get,

$$x_{i_j} - x_i = J(f(\cdot); \tau_i)(\tau_{i_j} - \tau_i) + e_{i_j} + n'_{i_j} \quad (3)$$
$$e_{i_j} \sim o(||\tau_{i_j} - \tau_i||_2)$$

where $x_{i_j}$ $(1 \leq j \leq m_i)$ are $m_i$ elements in the $\varepsilon$ ball of $x_i$ and noise vector $n'_{i_j}$ denotes $n_{i_j} - n_i$. By using matrix notation, we have,

$$X_i - x_i 1_{m_i}^T = J_i(\mathcal{T}_i - \tau_i 1_{m_i}^T) + E_i + N_i \quad (4)$$

Assume there are $m_i$ points $\{\tau_{i_j}, x_{i_j}\}_{j=1}^{m_i}$ in the $\varepsilon$-*neighborhood* of $\{\tau_i, x_i\}$, the local data matrix $X_i$ is defined as $[x_{i_1}, x_{i_2}, ..., x_{i_{m_i}}] \in \mathbb{R}^{D \times m_i}$. The corresponding latent co-ordinate matrix $\mathcal{T}_i$ for $X_i$ is denoted as $[\tau_{i_1}, \tau_{i_2}, ..., \tau_{i_{m_i}}] \in \mathbb{R}^{d \times m_i}$, the corresponding local Taylor approximation error matrix $E_i$ is denoted as $[e_{i_1}, e_{i_2}, ..., e_{i_{m_i}}] \in \mathbb{R}^{D \times m_i}$ and $N_i$ is the local inlier noise matrix $[n'_{i_1}, n'_{i_2}, ..., n'_{i_{m_i}}] \in \mathbb{R}^{D \times m_i}$. $J_i$ is the short notation of $J(f(\cdot); \tau_i)$.

It is natural to assume noise $n'_{i_j}$ to be i.i.d with homogeneous Gaussian distribution $N(0, \sigma_n^2 I_D)$ $(1 \leq j \leq m_i)$. Furthermore, we treat errors as independent Gaussian random vectors with different covariance matrices, i.e., $e_{i_j} \sim N(0, \sigma^2 I_D \alpha_{i_j})$. By adding noise and error together, an *integrated error* vector $\varepsilon_{i_j}$ is defined as,

$$\varepsilon_{i_j} = n'_{i_j} + e_{i_j} \sim N(0, \sigma_n^2 I_D + \sigma^2 \alpha_{i_j} I_D) \quad (5)$$
$$j = 1, 2, ..., m_i$$

where $\sigma_n^2$ indicates the scale of the noise's covariance and $\sigma^2$ indicates the scale of the error. $\alpha_{i_j}$ reflects the inhomogeneous property of different error of Taylor expansion on points $x_{i_j}$ (based on $x_i$). Thus, noise and error are jointly considered. Instead of treating error $e_{i_j}$ as homogeneous across different points, we argue the error is proportional to the relative distance $||\tau_{i_j} - \tau_i||_2$, based on the natural property of Taylor expansion.

**Inference.** To reflect the inhomogeneous property of error $\varepsilon_{i_j}$ (combine both Taylor approximation error and inlier noise), we propose the following objective function to estimate $J_i$,

$$\mathcal{L}(J_i) = ||E_i S||_F^2$$
$$= ||\{(X_i - x_i 1_{m_i}^T) - J_i(\mathcal{T}_i - \tau_i 1_{m_i}^T)\} S||_F^2 \quad (6)$$

where $S \in \mathbb{R}^{m_i \times m_i}$ is the diagonal weight matrix with $s_{jj} = s(\tau_{i_j}, \tau_i)$ indicating the importance to minimize error $\varepsilon_{i_j}$ on $x_{i_j}$. Intuitively, we emphasize the error minimization for $x_{i_j}$ more than $x_{i_k}$ if $s_{jj}$ is larger than $s_{kk}$.

Then we discuss how to choose $\alpha_{i_j}$ and determine $S$ accordingly. First, we choose $\alpha_{i_j} = \alpha(||\tau_{i_j} - \tau_i||_2)$ where $\alpha(\cdot)$ is a monotonically non-decreasing function in the non-negative domain. Supported by the fact of local-isometry (eq. 2), we have $\alpha_{i_j} \sim \alpha(||x_{i_j} - x_i||_2)$. So,

$$\varepsilon_{i_j} \sim N(0, (\sigma_n^2 + \sigma^2 \alpha(||x_{i_j} - x_i||_2)) I_D) \quad (7)$$
$$j = 1, 2, ..., m_i$$

Eq. 7 immediately suggests a weight function, $s_{jj} = s(||\tau_{i_j} - \tau_i||_2) = 1/(\sigma_n^2 + \sigma^2 \alpha(||x_{i_j} - x_i||_2))$. Given the function $\alpha(\cdot)$, $s(\cdot)$ can be determined and we get the weight matrix $S$. Then, eq.6 can be effectively solved by the following optimization framework

$$\arg\max_{J_i} ||(J_i^T \widetilde{X}_i S S^T \widetilde{X}_i^T J_i)||_*, s.t., J_i^T J_i = I_d \quad (8)$$

Essentially, the solution of eq. 8 is just the largest $d$ eigenvectors of the matrix $\widetilde{X}_i S S^T \widetilde{X}_i^T$ ($\widetilde{X} = (X_i - x_i 1_{m_i}^T)$), which is called *local structure matrix* $T_i \in \mathbb{R}^{D \times D}$. The local intrinsic dimensionality $d$ can be estimated by finding the largest gap between eigenvalues of $T_i$ (Mordohai & Medioni, 2010).

**Analysis.** By modeling inhomogeneous error $e_{i_j}$, curvature effect (Hessian) is implicitly considered without high order terms. Furthermore, for a large range of $\alpha(\cdot)$ (such as high order polynomial), weight $s_{jj}$ is quite small when $||x_{i_j} - x_i||_2$ is large. Thus, outliers will not affect the estimation results much.

It is also interesting to compare the local learning methods with different kernel functions $\alpha(\cdot)$. For standard low-rank matrix approximation by Singular Value Decomposition (SVD), $\alpha(\cdot)$ is a constant function. This is because SVD assumes data lie on a linear subspace without considering curvature or outliers. SVD can be improved to Robust SVD, by introducing the robust influence function to handle outliers (De la Torre & Black, 2003). However, Robust SVD is still constrained by the linear model and the computational cost is high because of the iterative computation. On the other hand, Tensor Voting (TV) uses Gaussian kernel (Mordohai & Medioni, 2010), which can be viewed as a special case of $s(\cdot)$ when $\sigma_n = 0$. This is because standard Tensor Voting does not consider inlier noise.

## 4. Global Manifold Structure Learning

In this global stage of RMMSL, we focus on multiple smooth manifolds clustering based on local structure learning results.

In contrast to previous works, the clustering stage in RMMSL can handle multiple non-linear (possibly intersecting) manifolds with different dimensionalities and it explicitly considers outlier filtering, which is addressed as one step in the clustering process. Compared to the standard assumptions in clustering, i.e., the intra-class distance should be small while the inter-class distance should be large, we further argue that each cluster should be a smooth manifold component. As shown in Fig. 2, when two manifolds intersect, there exist multiple possible clustering solutions, while only the rightmost is the result we want. The underlying assumption we make is local manifold has relatively low curvature, i.e. it changes smoothly and slowly as spatial distance increases. In order to get the *flattest* manifold clustering result, it is natural to incorporate a *flatness* measure into the clustering objective function.

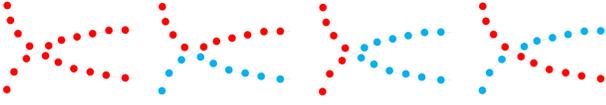

Figure 2. An example of multiple smooth manifolds clustering. The first one is the input data samples and the other three are possible clustering results. Only the rightmost is the result we want because the underlying manifolds are smooth.

**Curved-Level Measure.** As an approximation of (continuous) Laplacian-Bertrami operator (Hein et al., 2005), (discrete) graph Laplacian can measure the smoothness of the underlying manifold. However, computing graph Laplacian is a global process and most of the theoretical results can not be easily adapted to the multi-manifold setting. On the other side, curvature is a local measurement to indicate the curved degree of a geometric object. As a generalization of the curvature for 1D curve and 2D surface, the Ricci curvature tensor is proposed to represent the amount by which the volume element of a geodesic ball in a curved (Riemannian) manifold deviates from that of the standard ball in Euclidean space.

Inspired by the idea of curvature, the following *curved-level* measurement $R(x)$ is considered,

$$R(x) = \sum_{x_i \in N(x)} \frac{||\theta(J_i, J)||}{d(x_i, x)} \quad (9)$$

$\theta(J_i, J)$ measures the principal angle between the tangent space $J_i \in \Re^{D \times d_i}$ ($T_{x_i}\mathfrak{M}$) and $J \in \Re^{D \times d}$ ($T_x\mathfrak{M}$). $d(x_i, x)$ is the geodesic distance between $x_i$ and $x$. $N(x)$ is the spatial neighborhood points set for $x$. Intuitively, $R(x)$ is analogous (up to a constant variation) to the integration of the unsigned principal curvatures along different directions at $x$. The theoretical analysis on the connection between $R(x)$ and curvature (such as mean curvature) as well as the asymptotic behavior when the number of data samples goes to infinity is left for future investigation. Based on eq. 9, we can measure the (approximate) *total curved* level on one manifold cluster $\mathfrak{M}_k$ by summing up $R(\cdot)$ on all data samples $x_i$ belonging to this cluster,

$$R(\mathfrak{M}_k) = \sum_{x_i \in \mathfrak{M}_k} R(x_i) = \sum_{x_i \in \mathfrak{M}_k, (i,j) \in G} \frac{||\theta(J_i, J_j)||}{d(x_i, x_j)} \quad (10)$$

where $G$ is the undirected neighborhood graph built on data samples ($\varepsilon$-neighborhood or $K$-nearest neighborhood graph).

**Objective Function.** Eq. 10 provides an empirical way to measure the *total curved degree* on one particular manifold cluster $\mathfrak{M}_k$. This can be viewed as one type of the intra-cluster dissimilarity in the standard clustering framework. In order to get the balanced clustering results, we also consider the inter-cluster dissimilarity function as follows,

$$R(\mathfrak{M}_k, \mathfrak{M}_l) = \sum_{x_i \in \mathfrak{M}_k, x_j \in \mathfrak{M}_l, (i,j) \in G} \frac{||\theta(J_i, J_j)||}{d(x_i, x_j)} \quad (11)$$

In practice, the value of $||\theta(J_i, J)||/d(x_i, x)$ in eq. 9 is unbounded and numerically unstable. Thus, it is straightforward to compound eq. 9, 11 and the standard similarity kernel function (such as Gaussian) to get the *normalized* curved measurement. In particular, we use the standard minimization framework by putting $||\theta(J_i, J)||/d(x_i, x)$ into the similarity function with an additional distance similarity function,

$$J_{RMMSL}(\{\mathfrak{M}_k\}_{k=1}^{n_c}) = \sum_{k=1}^{n_c} \frac{W(\mathfrak{M}_k, \overline{\mathfrak{M}_k})}{W(\mathfrak{M}_k)} \quad (12)$$

where $W(\cdot)$ is the *contrary* version of the curved measure function $R(\cdot)$, i.e., the *flatter* the manifold is the *larger* value of $W(\cdot)$ is. $\overline{\mathfrak{M}_k}$ is the complementary set of $\mathfrak{M}_k$. The formulation of $W(\cdot)$ is,

$$\sum_{x_i \in \mathfrak{M}_k, (i,j) \in G} w_1\left(\frac{||\theta(J_i, J_j)||}{d(x_i, x_j)}\right) w_2(d(x_i, x_j)) \quad (13)$$

where $w_1(\cdot)$ and $w_2(\cdot)$ are similarity kernel functions that can be chosen as Gaussian or other standard formulations (similar for $W(\mathfrak{M}_k, \mathfrak{M}_l)$). Due to the shrinkage effect of kernel, $d(\cdot)$ is further approximated by Euclidean distance. Intuitively, $W(\mathfrak{M}_k)$ is one type of the intra-class similarity function on one manifold cluster and $W(\mathfrak{M}_k, \mathfrak{M}_l)$ is the inter-class similarity function between two clusters. The optimal $n_c$-classes clustering results are obtained by minimizing eq. 12.

**Algorithm.** Indeed, eq. 12 can be viewed as a multi-class normalized-cut with novel similarity measurement provided by local curved similarity and distance similarity functions (von Luxburg, 2007). Directly minimizing eq. 12 is an NP-hard problem, but it can be relaxed by graph spectral clustering in the following procedure.

*Step* 1. Before (global) manifold clustering, local structure learning of RMMSL in sec. 3 is performed to estimate the local tangent space $J_i \in \Re^{D \times d_i}$ at each point $x_i$. $d_i$ can be locally estimated from the local learning stage of RMMSL or chosen as a fixed value in advance. Also, the neighborhood graph $G$ is built on all input data samples $\{x_i\}_{i=1}^n$ ($n = n_1 + n_2$).

*Step* 2. Constructing the similarity matrix $W = [w(x_i, x_j)]_{i,j} \in \Re^{n \times n}$ by the following two kernels. The first kernel is the pairwise distance kernel, which is widely used in graph spectral clustering and defined as $w_1(x_i, x_j) = \exp\{-||x_i - x_j||^2/(\sigma_i \sigma_j)\}$. In particular, we use the idea from self-tuning spectral clustering to select the local bandwidth $\sigma_i$ and $\sigma_j$ (Zelnik-Manor & Perona, 2005). The second one is curved level kernel $w_2(x_i, x_j) = \exp\{-(\theta(J_i, J_j))^2/(||x_i - x_j||^2(\sigma_c^2/\sigma_i \sigma_j))\}$, where $\sigma_c$ is used to control the effect of this curved similarity. Then, $w(x_i, x_j)$ is set as

$$w_{ij} = w_1(x_i, x_j) w_2(x_i, x_j) = \exp\{-(\frac{||x_i - x_j||^2}{\sigma_i \sigma_j} + \frac{\theta(J_i, J_j)^2}{||x_i - x_j||^2 \sigma_c^2/\sigma_i \sigma_j})\}. \quad (14)$$

*Step* $\widetilde{3}$ *(optional).* Based on $W$, outlier detection (filtering) can be done as described later.

*Step* 4. Once we have the similarity matrix $W$, the standard spectral clustering technique can be applied. Specifically, we compute the (unnormalized) Laplacian matrix $L = D - W$, where $D$ is a diagonal matrix whose elements equal to the sum of $W$'s corresponding rows. We select the first $n_c$ (number of clusters) eigenvectors of the generalized eigenproblem $Le = \lambda De$. Finally, K-means algorithm is applied on the rows of these eigenvectors. After identifying manifold cluster labels, many tasks such as embedding and denoising can be performed on each manifold cluster.

**Outlier Detection.** We formulate the outlier detection problem as a manifold saliency ranking problem by the random walk model proposed in (Zhou et al., 2004). We define a random walk graph on $\{x_i\}_{i=1}^N$ from the following transition probability matrix $P = I - L_{rw} = D^{-1}W$. It can be shown that, if $W$ is symmetric, then the stationary probability $\pi$ of this random walk can be calculated directly without complex eigen-decomposition (Zhou et al., 2004),

$$\pi = 1_n^T D / ||W||_1 \quad (15)$$

where $1_n$ is an $n \times 1$ column vector with all elements as 1 and $||\cdot||_1$ is the entry-wise 1-norm of a matrix. It is straightforward that, ranking data according to $\pi$ is the same as the un-normalized distribution $1_n^T D$. After ranking, bottom points can be filtered out as outliers. The ratio of the outliers can be given as a prior or be estimated by performing K-means on $1_n^T D$.

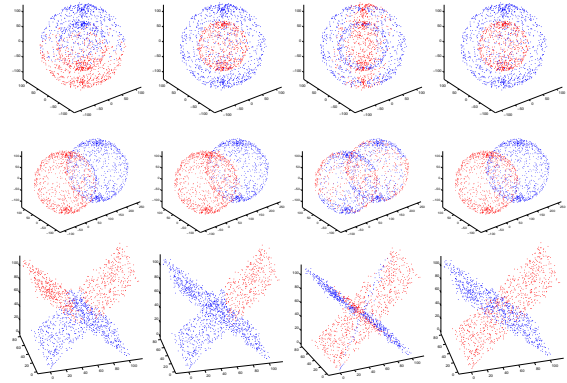

*Figure 3.* Visualization of part of the clustering results in Table 1. The first row : one noisy sphere inside another noisy sphere in $\Re^3$. The second row: two intersecting noisy spheres in $\Re^3$. The third row: two intersecting noisy planes in $\Re^3$. For each part from left to right: K-means, self-tuning spectral clustering (Zelnik-Manor & Perona, 2005), Generalized PCA (Vidal et al., 2010) and RMMSL.

**Analysis.** The key idea of this algorithm is to present a novel way to construct the similarity matrix $W$ in a multiple kernel setting (based on local structure estimations) to encourage *flatter* clustering results. It is worth noting that if two points have different tangent spaces, the similarity (eq. 14) becomes smaller when two points get closer (in a range). This can be viewed as an intuitive explanation that why RMMSL can handle multiple intersecting manifolds. From the high level point of view, the pairwise similarity incorporates the information on two local points sets rather than two single points only. Similar ideas were proposed in (Elhamifar & Vidal, 2009; Goldberg et al., 2009; von Luxburg et al., 2011).

## 5. Experiments

We evaluate the performance of RMMSL on synthetic data, USPS digits, CMU Motion Capture data (MoCap) and Motorbike videos. These data are chosen to demonstrate the general capability of RMMSL as well as the advantages on nonlinear manifolds. We investigate the performance of manifold clustering and further applications such as human action segmentation and motion flow modeling. We also perform quantitative comparisons of the local tangent space estimations. In particular, the weighted low-rank matrix decomposition (local structure learning in RMMSL) is compared with the local SVD (or local PCA (Teh & Roweis, 2003)) and ND-TV (Mordohai & Medioni, 2010). Results show that RMMSL is more robust to curvature and outliers than other methods. Also, if the manifold has relatively low curvature and is outlier free, then RMMSL and local SVD have similar results. Details are omitted due to space limit.

| Data/Methods | K-means | NJW Clustering | Self-tuning Clustering | GPCA | SSC | RMMSL |
|---|---|---|---|---|---|---|
| Big-small spheres | 0.50 | 1.00 | 1.00 | 0.51 | 0.56 | 1.00 |
| Two-intersecting spheres | 0.76 | 0.78 | 0.84 | 0.50 | 0.53 | 0.95 |
| Two-intersecting planes | 0.51 | 0.60 | 0.62 | 0.85 | 0.93 | 0.95 |
| USPS-2200 | 0.80 | 0.89 | 0.89 | - | - | 0.90 |
| USPS-5500 | 0.78 | 0.88 | 0.88 | - | - | 0.88 |
| CMU MoCap | 0.69 | 0.81 | 0.81 | - | - | 0.89 |
| Motorbike Video | 0.72 | 0.84 | 0.85 | 0.85 | 0.87 | 0.96 |

Table 1. Rand index scores of clustering on synthetic data, USPS digits, CMU MoCap sequences and Motorbike videos.

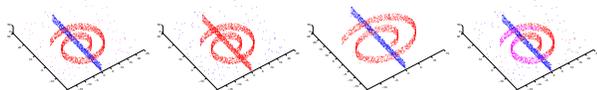

Figure 4. Clustering results of RMMSL on two manifolds with outliers. From left to right: ground truth, outlier detection, clustering after outlier filtering and clustering without outlier filtering.

### 5.1. Multiple Manifolds Clustering

In this task, quantitative comparisons of manifolds clustering are provided on three data sets. We compare RMMSL (global structure learning) to K-means, spectral clustering (NJW algorithm (Ng et al., 2002)), Self-tuning spectral clustering (Zelnik-Manor & Perona, 2005), Generalized PCA (GPCA) (Vidal et al., 2010) and Sparse Subspace Clustering (SSC) (Elhamifar & Vidal, 2009). These methods (except K-means) are chosen because they are related to manifold learning or subspace learning. We also perform comparisons with other spectral clustering and subspace clustering methods such as (Yan & Pollefeys, 2006) and (Goh & Vidal, 2007), and get the same conclusion, but results are omitted due to space limit. Rand Index score is used as the evaluation metric. The kernel bandwidth $\sigma$ in spectral clustering is tuned in $\{1, 5, 10, 20, 50, 100, 200\}$ for synthetic data, MoCap and videos, and $\{100, 500, 1000, 2000, 5000\}$ for USPS. The value of the $K_{th}$ neighborhood in self-tuning spectral clustering and RMMSL is chosen from $\{5, 10, 15, 20, 30, 50, 100\}$. $\sigma_c$ in the global stage of RMMSL is chosen from $\{0.2, 0.5, 1, 1.5, 2\}$. In the local stage, quadratic $\alpha(\cdot)$ is used and $\sigma_n$ is set as 1. The sparse regularization parameter of SSC is tuned in $\{0.001, 0.002, 0.005, 0.01, 0.1\}$. For synthetic data with random noise, the parameters for all methods are selected on 5 trials and then the average performance on another 50 trials is reported. For real data, parameters are selected by picking the best Rand Index. For all methods containing K-means, 100 replicates are performed.

**Synthetic Data.** Since most clustering algorithms do not consider outliers explicitly, we first perform a comparison on 3 outlier free synthetic data, while each contains 2000 noisy samples from two manifolds in $\Re^3$ ($d = 2$ and $D = 3$). Results are shown in Fig. 3. Rand index scores are given in the first three rows of Table 1. For all methods, the number of clusters $n_c$ is fixed as 2. Results (Table 1) show that RMMSL achieves comparable and often superior performance than other candidate methods. In particular, when two manifolds are nonlinear and have intersections, such as two intersecting spheres (second row), the advantage of RMMSL is clearest.

To verify the robustness, we further evaluate RMMSL on synthetic data with outliers. We add 100 outliers and a 2D plane (1000 samples) on the Swiss roll (2000 samples) to generate two intersecting manifolds in $\Re^3$. The results of RMMSL are shown in Fig. 4. RMMSL effectively filters out outliers and achieves 0.96 Rand score ($n_c = 2$) and 0.99 F-measure for outlier detection when the ratio is given. Also, the Rand score is reduced to 0.78 if we do clustering without outlier filtering ($n_c = 3$). This fact suggests that the outlier detection step is helpful if outliers exist. It is worth noting that spectral clustering methods cover broader cases than RMMSL, which mainly has advantage on multiple low-dimensional manifolds embedded in high dimensional space. For instance, in the case of two 2D Gaussian distributed clusters in $\Re^2$, RMMSL is reduced to self-tuning spectral clustering (all local tangent spaces are ideally identical). Compared with multiple subspace learning methods such as GPCA and SSC, which are the state-of-the-art for *linear* manifold clustering, our approach is better when underlying manifolds are *nonlinear*.

**USPS Digits.** We choose two subsets of USPS hand written digits images. The first contains 1100 samples for digits 1 and 2 each (USPS-2200) and the second contains 1100 samples for digits 1 to 5 each (USPS-5500). $D$ is reduced from 256 (size $16 \times 16$ images) to 50 by PCA and $d$ is fixed as 5. Due to the highly nonlinear image structure, results of subspace clustering methods are not reported. For USPS data, the possible high intrinsic dimensionality v.s. the limited number of samples bring difficulties for data structure learning, especially the local learning stage of RMMSL. Nevertheless, RMMSL achieves comparable results.

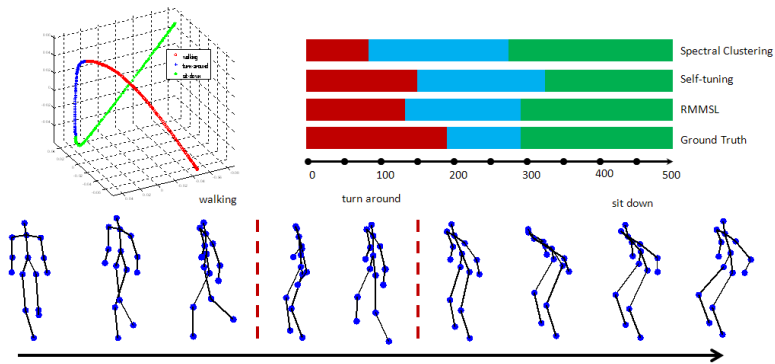

*Figure 5.* An example of human action segmentation results on CMU MoCap. Top left, the 3D visualization of the sequence. Top right, labeled ground truth and clustering results comparison. Bottom, 9 uniformly sampled human poses.

**MoCap Data.** The automatic clustering of human motion sequences into different *action units* is a necessary step for many tasks such as action recognition and video annotation. Usually this is referred as *temporal segmentation*. In order to make a fair comparison among different clustering methods, we focus on the non-temporal setting, i.e., the temporal index is removed and sequences are treated as collections of static human poses. We choose 5 mixed action sequences from subject 86 in the CMU MoCap. We use the joint-position (45-dimensional representation for 15 human body markers in $\Re^3$) features which are centralized to remove the global motion. The average Rand scores are reported in Table 1. It shows that RMMSL achieves higher clustering accuracy than other candidate methods.

One motion sequence (500 frames) and the corresponding results are visualized in Fig. 5. The subject walks, then slightly turns around and sits down. By combining the local learning results from RMMSL and (Teh & Roweis, 2003), joint-position features ($\Re^{45}$) from this sequence are visualized in $\Re^3$. This figure supports the assumption that there are low-dimensional manifolds in the joint-position space. In fact, this MoCap sequence can be viewed as three *connected* nonlinear motion manifolds, corresponding to walking, turn-around and sit-down respectively.

**Motion Flow Modeling.** Unsupervised motion flow modeling is performed on videos (Lin et al., 2011). The goal is to analyze coordinated movements formed by multiple objects, extract *semantic* level information, and understand what's happening in the scene. Given motorbike videos as shown in Fig. 6 (from YouTube), global motion pattern is learned from low level motion features. This is an important task for video analysis and can be served as a foundation for many applications such as object tracking and abnormal event detection (Lin et al., 2010; 2011).

Differ from the probabilistic approaches in (Lin et al., 2011), we formulate motion flow learning as a manifold clustering problem. In the experiments, optical flows on salient feature points are estimated by Lucas-Kanade algorithm. Every feature point has 4D information of $(x, y, v_x, v_y)$. Then motion direction $\theta$ is calculated, and every point is embedded to $(x, y, \theta)$ space. We observe that coordinated group movements form into manifold structures in $(x, y, \theta)$ space. Therefore, the points are used as the input. The first video contains $n = 9266$ points and the second one contains $n = 8684$ points. We use RMMSL to learn the global motion manifolds by doing manifold clustering ($n_c = 2$) and get the best Rand scores which are reported in Table 1. Since optical flow results are noisy, outlier filtering is performed before clustering. The motion manifold learning results of two motorbike videos are shown in Fig. 6, where motion manifolds are visualized on images after kernel density interpolation. From the results we can see that the clustered manifolds have clear semantic meanings, since each manifold corresponds to a coordinated movement formed by a group of motorbikes. Therefore, RMMSL correctly learns global motion to help understand the video scenes.

## 6. Conclusion

Robust Multiple Manifolds Structure Learning is proposed to effectively learn the data structure by considering noise, curved-level and multiple manifolds assumption. In particular, the estimated local structure is used to assist the global structure learning tasks of clustering and outlier detection. The algorithm is evaluated and compared to other state-of-the-art clustering methods. The results are encouraging: on both synthetic and real data, the proposed algorithm yields smaller clustering errors in challenging cases, especially when multiple nonlinear manifolds intersect. Furthermore, the results on action segmentation and motion flow modeling demonstrate RMMSL's capability for broad challenging applications in real world.

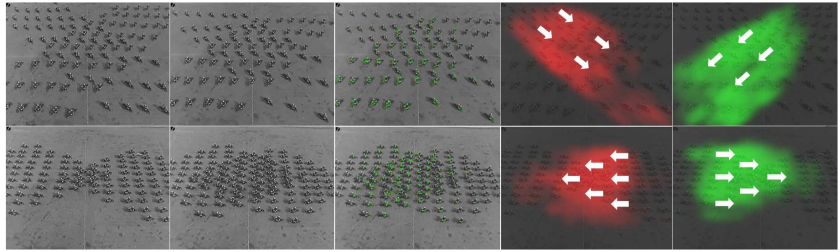

*Figure 6.* Two examples of motion flow modeling results on motorbike videos. From left to right: two images samples, optical flow results, learned motion manifolds with highlighted motion directions.

## Acknowledgements

This work was supported in part by NIH Grant EY016093 and DE-FG52-08NA28775 from the U.S. Department of Energy. We thank Fei Sha for helpful discussions.